\documentclass[runningheads]{llncs}

\usepackage{eccv}

\usepackage{eccvabbrv}
\usepackage{xcolor}
\usepackage{graphicx}
\usepackage{booktabs}

\usepackage[accsupp]{axessibility}  

\usepackage{hyperref}

\usepackage{orcidlink}

\begin{document}

\title{V-REX:\\
Efficient Specialist VLM Training for Veterinary X-Rays} 

\titlerunning{Abbreviated paper title}

\author{Tim Elsner\inst{1}\orcidlink{0000-0003-1309-8003} \and
Nicole McNally\inst{2}\orcidlink{0009-0009-0112-9758} \and
Andre Dourson\inst{2}\orcidlink{0009-0007-1721-6472} \and
Michael Fitzke\inst{2}\orcidlink{0009-0009-5849-1336}}

\authorrunning{T.~Elsner et al.}

\institute{Vyyo AI \and
Mars Petcare\\
\email{nicole.mcnally@effem.com}}

\maketitle


\begin{abstract}
While generalist VLMs are expensive to train, creating domain experts is widely assumed to require fine-tuning increasingly large foundation models. We show that, in veterinary radiology, this assumption is misguided. By rethinking the entire VLM pipeline - from text tokenisation and pre-training to grounding and inference - we demonstrate that careful engineering can yield models that outperform much larger foundation models from scratch, without relying on any other data. Our approach introduces new strategies for generative pre-training and grounding that improve training efficiency, increasing data utilisation and downstream performance. Using only a fraction of the parameters, data, and compute of contemporary generalist models, we develop the first VLM capable of generating diagnostic reports for veterinary radiographs, surpassing open foundation models on this task by significant margin.
\end{abstract}

\section{Introduction}
Large Language Models \cite{fewshot, deepseek, llama} (\textit{LLMs}) have demonstrated remarkable capabilities across a wide range of natural language tasks that have found their way into everyday life, including reasoning, question answering, and few-shot generalisation. In parallel, vision models have even surpassed human performance for tasks like visual recognition \cite{faces, liu2025human} or detection \cite{Browatzki2024-a, wang2023yolov7}. Vision-Language Models (\textit{VLMs}) bring both domains together, producing a text output from visual input data through using an open vocabulary. These models typically extend an existing, pre-trained LLM backbone, fine-tuning it on how to shift the output distribution according to a visual input signal. While powerful, this design makes such systems highly expensive in terms of compute, limiting their possible training and application for many purposes: Such models have a much more vast pool of language and knowledge than what needed for a small, reliable expert model.\\
Such generalist or \textit{foundation model VLMs} are jacks of all trades, suited for a wide range of daily tasks, but also masters of none: A model used for radiology that is also able to identify tropical plants while also being able to transcribe handwriting and labelling cars may not be able to outperform an experienced radiologist.\\~\\
The general VLM pipeline can be broken down into the following parts that have a number of caveats with the currently dominating, jack of all trades approach:\\
\textbf{Tokenisation}~ A VLM doing radiology in English language does not need snippets from coding or Chinese characters. A general tokenisation inflates sequence length, meaning training and generation are more expensive and error-prone: More chances for mistakes exist if the output gets longer while attention becomes more costly and imprecise \cite{dilution}.\\
\textbf{Pre-Training}~ 
A pre-trained vision module that already knows how to extract features can be helpful, but is often too general for an expert task: If only $0.01 \%$ of the space is used to \eg represent x-rays, building a dedicated unsupervised approach for x-ray can be the better choice. Similarly for language, we do not need taco recipes or French when writing English x-ray reports.\\
\textbf{Training}~ Even simply fine-tuning larger models can be too expensive. An already distilled (and hence, heavily maimed) model at \eg $3B$ parameters is still expensive to train. Even when applying techniques like LoRA \cite{lora}, we effectively try to find a low-rank adjustment of a space where only the tiniest part is relevant. Reacquiring that knowledge with a small model can be much faster than re-purposing a huge model where only a fragment of the learned knowledge is useful: Rather than combing through tons of hay, it's more convenient to just get a new needle.\\
\textbf{Inference}~ Especially with very limited training data, the already produced text alone can be too strong of a prior. This can lead to the image input eventually being ignored. Produced outputs are also rarely explainable and strategies like multi-image VLMs need to be included in training, all leading to higher costs.\\~\\
We use veterinary x-ray data as an exemplary task to show how to improve each step in the VLM pipeline, with medium-sized proprietary data ($15m$ text-image-pairs) and medium-sized compute (an individual H100 for most experiments). Yet our approach produces model that beats generalist behemoths at a fraction of their costs, trained from scratch, with a lightweight codebase only consisting of Pytorch \cite{torch} and Lightning \cite{lightning}.\\
We aim for reproducibility, effective data and compute usage, and ease of implementation, delivering a framework that works well for specialised VLM tasks, referred to as V-REX. \textbf{Our core contributions are as follows:}
\begin{enumerate}
    \item We offer a best practise recipe to train a VLM from scratch, on a budget, with experiments to help estimate the required amounts of data and compute to beat large foundation models
    \item We provide a new pre-training strategy that makes training cheaper, but also pushes the achievable boundaries by being more data efficient
    \item We contribute strategies for inference that allow \eg multi-image processing and better image \textit{grounding}, reducing language priors in favour of image content
\end{enumerate}

\section{Related Work}\label{app:related_work}
We follow the common definition of VLMs as any text-producing model that generates text according to a visual input \cite{vlm}, most commonly images. More precisely, captioning VLMs (like our proposed model) produce one text for a given visual input. Our work aims to make VLMs more efficient both in terms of compute and required data. Hence in our case, while training VLMs from scratch is possible for both CNN and Transformer-based architectures \cite{vilbert, cnn_vlm_from_scratch}, using a pre-trained vision encoder has proven to be more effective. 

We thus discuss VLMs in general (\cref{sec:vlm}), then focus on contrastive embeddings for improved performance (\cref{sec:contrastive}), and a specific contrastive embedding called RAPTOR used for our work as pre-trained vision encoder.

\subsection{VLMs}\label{sec:vlm}
Early work on vision-language modelling focused on directly mapping images to text via encoder–decoder architectures \cite{showandtell}, \eg combining a CNN image encoder with an LSTM \cite{lstm} decoder to generate captions. This line of work established the now-standard paradigm of extracting a fixed visual representation and autoregressively generating text conditioned on it. This encoder-decoder design of captioning VLMs follows the transformer \cite{transformer} structure of an encoder for the image and an autoregressive decoder for the text \cite{flamingo}.\\
Dual-stream architectures such as \cite{vilbert} and \cite{xlmert} introduced separate visual and textual encoders with cross-attention layers to enable fine-grained alignment between modalities. These models were often pre-trained with multiple objectives, including masked language modelling, masked region modelling, and cross-modal matching.\\
More intricate VLM paradigms consist of two independent encoders, one for text and one for visuals, often also trained with an additional contrastive embedding. Approaches like the PaliGemma family \cite{paligemma, paligemma2} follow this design. For both approaches, variants exist that utilise existing language models to ease the computational burden \cite{zhu2023minigpt, llava, dai2023instructblip}.\\
Note that for all the mentioned models, our proposed VLM is orders of magnitude smaller ($800m$ parameters for the largest and uncompressed V-REX variant compared to \eg $7B$ for the smallest and already distilled LLaVA \cite{llava}), while general VLMs also require orders of magnitudes more data than is available for most scenarios. To improve these the data- and computation cost hurdles of VLM training, we utilise tokenisation and pre-training in particular, as both offer ways to significantly reduce costa while also improving performance.\\~\\
\textbf{Tokenisation} Just as with pure LLMs, performance of text-producing networks directly depends on how text is captured. While some approaches skip tokenisation entirely \cite{aleph_alpha_tokeniser,meta_tokeniser}, most approaches still rely on Byte Pair Encoding (recursively replacing the most frequent pair of tokens with a single new token) \cite{bpe}, even for other domains like images \cite{mdbpe}, or apply related strategies like WordPiece \cite{bert} or SentencePiece \cite{sentencepiece} that build a vocabulary according to heuristics. A domain-specific vocabulary can hence drastically reduce the average sequence length, reducing computational complexity while allowing less room for mistakes.\\~\\
\textbf{Pre-Training} As VLMs often utilise already trained language models \cite{zhu2023minigpt, llava, dai2023instructblip} or vision encoders \cite{vilbert, xlmert}, this can be considered as a form of pre-training. However, for our purposes, we also consider approaches that do generative pre-training on visual data. Chen \etal \cite{chen2020generative} learn to autoregressively predict pixels to learn a rich representation that can be used for other tasks like classification. Similar in spirit, Xie \etal \cite{xie2023visorgpt} does the same for a decoder-only transformer in GPT-style, while others \cite{bao2021beit} utilise BERT-Style \cite{bert} masked image modelling to learn semantic features. Our method shows the effectiveness of the same principals to VLMs, also showing effectiveness on already rich representations.

\subsection{Contrastive Embeddings}\label{sec:contrastive}
\textbf{Explicit} contrastive embedding methods like CLIP \cite{clip} and related approaches \cite{yu2022coca, jia2021scaling, siglip} produce an output vector that describes a given input text or image, shaped such that the dot product between visual and text inputs measures the similarity. This is obtained through a contrastive loss that usually utilises similar (positive) and dissimilar (negative) inputs. \textbf{Implicit} contrastive embedding methods like DINO \cite{dino} and similar approaches \cite{grill2020bootstrap,chen2021exploring} forgo the contrastive embedding, instead \eg relying on regularising the network enough to avoid the necessity of negative pairs and producing features in an unsupervised manner.

\begin{figure}[ht]
  \begin{center}
    \centerline{\includegraphics[width=0.7\columnwidth]{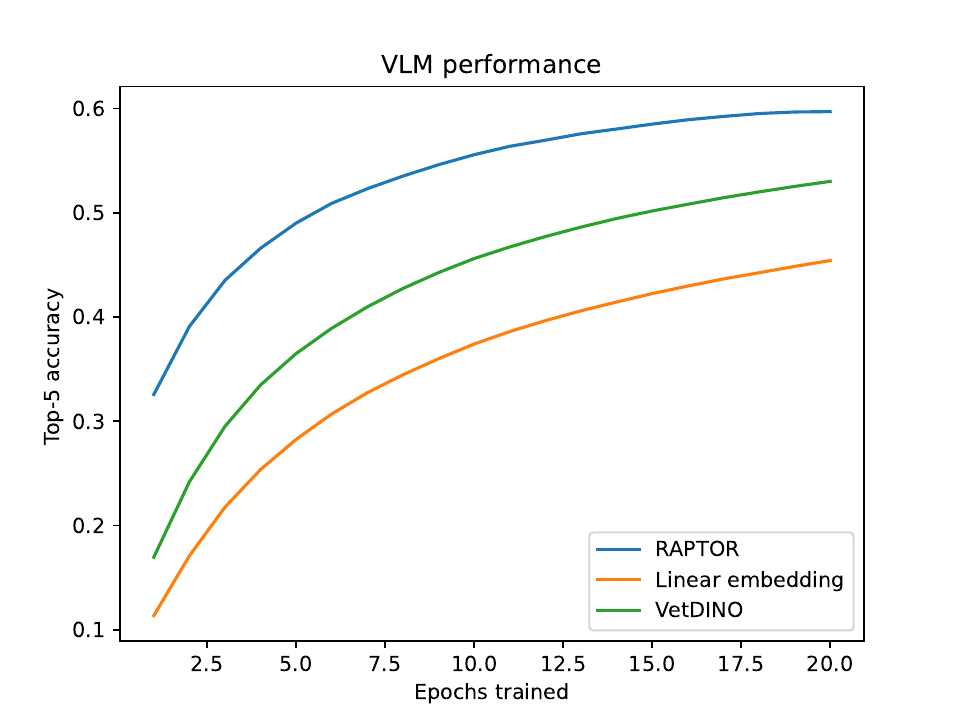}}
    \caption{
      Advantage of using RAPTOR: With the right pre-trained data embedding, our V-REX performs much better, however, still trains with just a linear embedding.
    }
    \label{fig:raptor}
  \end{center}
\end{figure}

We can bring together the advantages of both worlds, using unsupervised and rich features that are steered towards more semantically meaningful results, using the text as a guidance for visual features towards what is relevant. We employ these \textbf{RAPTOR embeddings} as the backbone for our VLM, produced by a (currently) proprietary encoder trained on veterinary data. RAPTOR uses the DINO \cite{dino} architecture with a contrastive, CLIP-like \cite{clip} objective between embedded images and text, built upon Vet-DINO \cite{vetdino}. We use the feature tokens of the last layer of the transformer, right before the extraction of the CLS token. Essentially, we use RAPTOR to get an already rich visual representation of the image with features that are more relatable to text tasks. While our model also works with DINO features or just image projections, we observe better optima and faster convergence using RAPTOR tokens (see \cref{fig:raptor}).

\begin{figure*}[ht]
  \vskip 0.2in
  \begin{center}
  \centerline{\includegraphics[width=1\columnwidth]{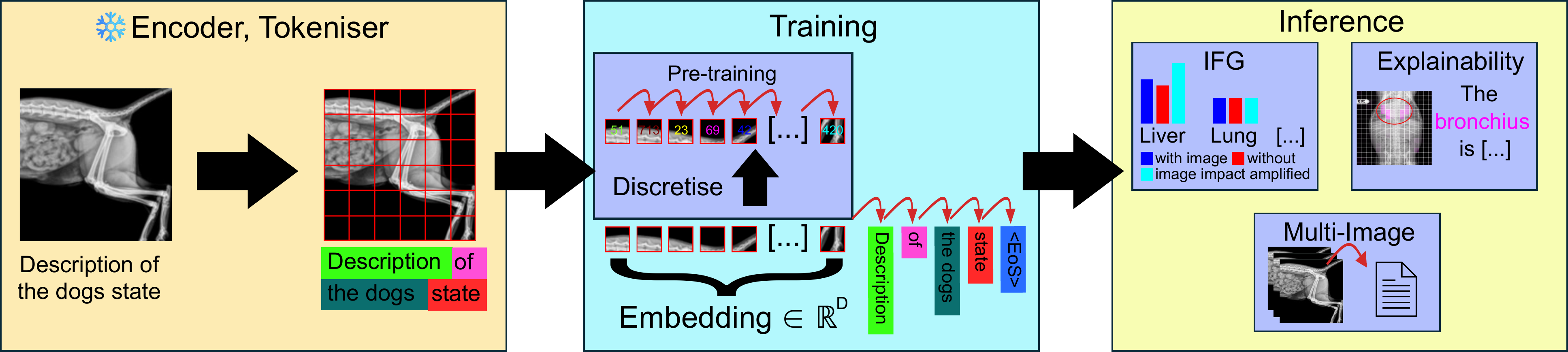}}
    \caption{Our proposed pipeline follows a fully autoregressive and unified architecture: For the encoded, frozen embeddings of the image, autoregressively learn to predict the next (quantised) image token as a pre-training. We then autoregressively learn to produce text. All of our model is using the same space, with a simple sequence with image tokens row by row followed by the text. Extensions like multi-image support or IFG for better image adherence are applied during inference.}\label{fig:architecture}
  \end{center}
\end{figure*}

\section{A Scaleable VLM Pipeline}
Unlike most approaches \cite{zhu2023minigpt, llava, dai2023instructblip} that train VLMs, we propose to approach each step of the "traditional" VLM pipeline from scratch for our target  \textit{expert domain} (here, veterinary data).\\
Commonly, VLMs rely on existing LLMs that are fine-tuned to also process images \cite{paligemma, flamingo}, hence keeping the general LLM tokeniser instead of using an expert vocabulary. We argue that while this eases the required computational burden by utilising  existing language and expert knowledge, it also comes at the cost of a non-expert tokenisation. Our veterinary model does not need tokens for programming and mandarin. To see the effect of tokenisation, consider the error probability of a sequence:
\begin{equation}
    P(error) = 1 - (1 - e)^T
\end{equation}
for $T$ tokens and an error probability per token of $e$. For an error rate per token of only $e = 0.02$ and a sequence length of $T = 20$, reducing the sequence length to $T = 10$ increases the chance of generation without a faulty step from about $66 \%$ to $81 \%$. We take this as an inspiration to revisit the entire VLM pipeline, in particular tokenisation, \textit{from scratch} while keeping the used components simple and lean, avoiding \eg upgrades to the transformer architecture \cite{deepseek_residual} or flashy new tokenisers \cite{meta_tokeniser, aleph_alpha_tokeniser}. We use a schedule-free \cite{schedulefree}, rectified \cite{radam} version of Adam \cite{adam}, forgoing any need for learning rate scheduling. 
Because there is a merit not only in explaining what worked, but also in what did \textit{not} work, we list our failed experiments in Appendix \ref{app:unsuccessful}.

\subsection{Text Preparation and Tokenisation}
Tokenisation schemes like BPE \cite{bpe} compress text according to heuristics, and even out the information density: Highly frequent pairs with low entropy of the input distribution are compacted together. This general strategy of applying generation to a more compact domain has been crucial for generative quality, also in other domains \cite{latentgan, latentdiffusion}. However, when switching from a general purpose to an expert domain VLM, tokenisation is often left as is. We first homogenise text through converting to lower case and removing typos with an LLM. We then apply a simple approach of first using a word-level tokeniser, then apply BPE on the word-level until we reach a vocabulary of $32768$ tokens. This gives us a high degree of compression with tokens that are still well interpretable (as the initial tokens are on word-level). For our data, this reduces the average text length from about $520$ characters to $30.8$ tokens (about $16.9$ characters per token), compared to an average of $106$ tokens (about $4.9$ characters per token) for the GPT 5.x tokeniser (roughly $6$ times longer sequences).

\subsection{VLM Architecture}
As the core of our model, we use a standard decoder-only transformer \cite{transformer} that learns to autoregressively predict the text part of a sequence consisting of image tokens followed by the report (see \cref{fig:architecture}). While flat projection of image patches did work, for our veterinary image tokens, we use RAPTOR-tokens (a DINO-CLIP hybrid trained on our same data, see \cref{sec:contrastive}).\\
As a pre-training step, we autoregressively train on the discretised image embeddings. For the text prediction training, we then allow full visibility within the image tokens, \ie set the upper-left part of the mask that corresponds to the image tokens to $0$. For the image tokens, we use a standard 2D sinusoidal encoding to annotate the position of the image embeddings, while applying the 1D sinusoidal encoding for the text position. We additionally add one embedding to all image tokens and one embedding to all text tokens, easing the distinction between image and text. We forgo more complicated (and powerful) encoding schemes \cite{rope, nope, drope}, as our reports largely follow the same structure, are relatively short, and for simplicity. We also use the same hyper parameters for the different phases of training. While we do not discuss these changes here, our architecture can be also trained to work on multi-image inputs, simply concatenating the tokens of multiple images.

\subsection{Optional Pre-Training}
We apply generative pre-training on our images for two particular reasons:
\textbf{First}, it enables the use of potentially unlabeled data, such as images without accompanying text. \textbf{Second}, the increased data utilisation allows for faster convergence to a better optimum.
We argue that generative pre-training is the most wholistic form of learning, as learning to replicate an input data distribution requires understanding of the fabric of the data distribution. Training to autoregressively predict the next part of the image allows the transformer to already understand the structure and inner workings of the images.~\\
In our case, we do all our experiments on already embedded $16-$by$-16$ image patches of the RAPTOR framework.
To enable autoregressive prediction of continuous values, we simply cluster the produced embeddings into 512 centres to obtain a discrete index representing groups of similar embeddings. As the assignment of a k-means cluster centre becomes memory intensive for high dimensional spaces, we obtain $k$ different indices representing the different image patches by rounding the vectors representing the image patches into $k$ different buckets. We do this efficiently by applying a binary quantisation on a projected subspace of the embedding vectors, with details on this given in Appendix \ref{app:bin_subspace_quant}.\\
During pre-training, we only use the quantised versions of the embeddings to get a target for prediction, \ie use the non-quantised version as input sequence to align with the later steps of using the unquantised image tokens.
\subsection{Training}
When continuing from the pre-training, we first train the head for $1500$ steps, before gradually warming up the learning rate over the next $1500$ steps. We randomly zero out the image embedding for $1$ in $32$ tokens in order to later identify and amplify the impact of the image on the text token prediction, achieving better grounding (see \cref{sec:ifg}).

\subsection{Inference-Time Improvements}\label{sec:ifg}
While specifically training models to deliver more conceivable results or utilising multiple images is possible, smaller projects often are limited in terms of compute or data. We hence provide a small set of suitable extensions.
\paragraph{Overcoming the language prior: Visual Grounding}
The already produced language can be a powerful prior for the next token predictor of VLMs, ignoring possible image content \cite{mirage}. We adapt the idea of classifier-free guidance from Ho \etal \cite{ddpm} that propose to arithmetically increase the impact of the controlling signal. For this, they compute and amplify the difference in output with and without the controlling signal (in their case: text).\\
In our case, for each next token step, we call the model twice, with and without image, measure the probabilities for the next tokens, and then add the difference multiplied with a scalar:
We define the next token probabilities given the previous $n$ text tokens $t$ and image $I$ as $P(t, I)$, with $t = (t_0, t_1, .., t_n)$. We then define our \emph{image-free guidance} (IFG), in analogy to classifier-free guidance in diffusion models \cite{ddpm}, with a magnitude factor $s$ as:
\begin{equation}
P_{IFG}(t, I) = P(t, I) + s\cdot (P(t, I) - p(t, \emptyset))
\end{equation}
This amplifies the contribution of the part of the signal that is specifically resulting from the image, while leaving tokens that result from the language prior alone unchanged. We then normalise the probabilities and sample the next token.

\paragraph{Multi-Image Prediction}
While we can train our model to use multiple images, we can just pool the results of predicting the next token from multiple images, always taking the most likely token over all images. This can increase accuracy for multi-image examples by up to $0.03$ in top-1 accuracy without any changes to the model.

\begin{figure}[t]
    \centering
    \includegraphics[width=0.6\linewidth]{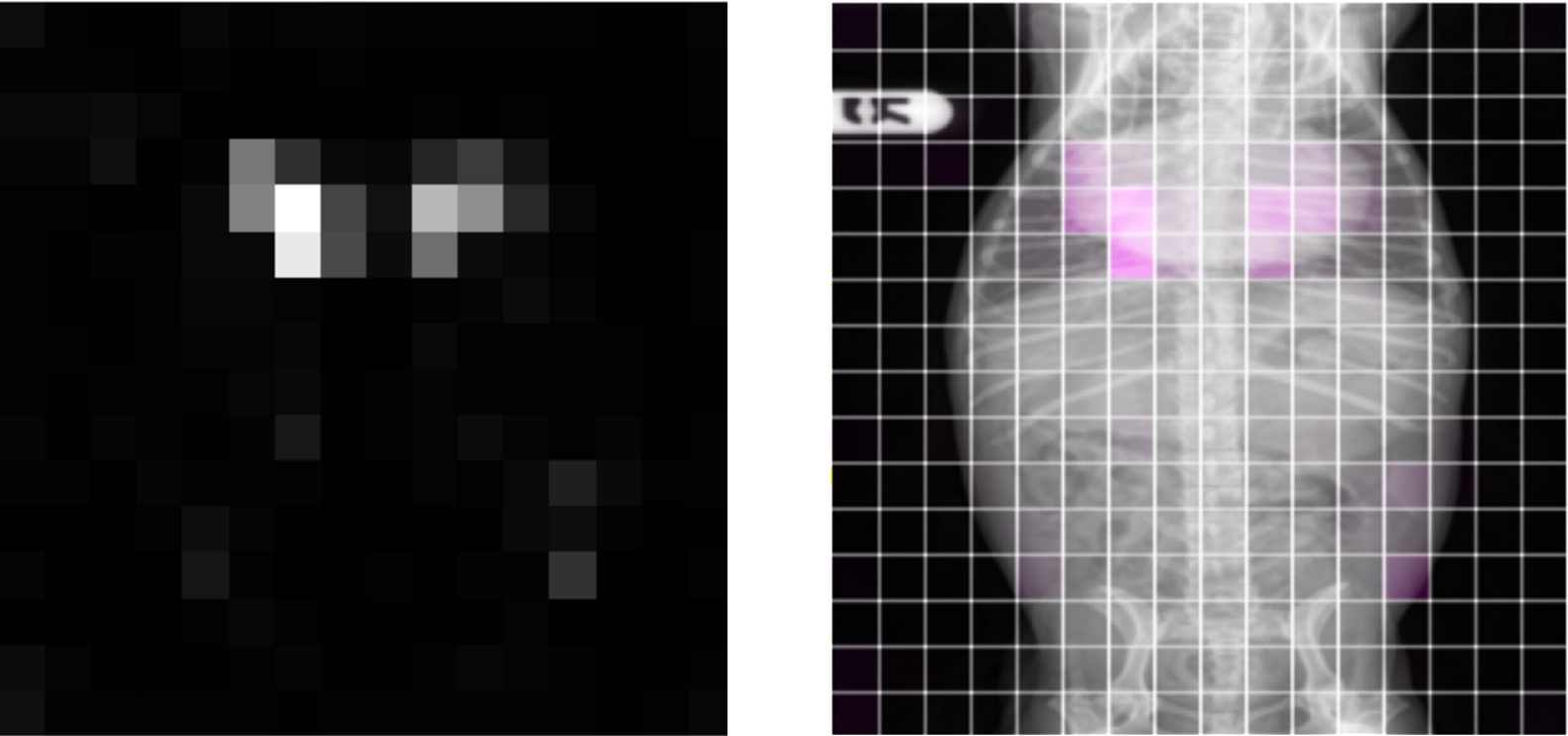}
    \caption{Visualisation of visual token impact: The model decides for a next token piece containing the word "cardiac", with the computed Integrated Gradients highlighting the heart.}
    \label{fig:explainability}
\end{figure}

\paragraph{Explainability}
As the used RAPTOR image tokens (opposing to a vanilla DINO) have noisy attention maps through the additional grounding in text that can not be recognised directly anymore, we can apply the \textit{Integrated Gradients} method \cite{IG} to show the visual attribution per text token (see \cref{fig:explainability}).

\section{Evaluation}
%
%



We split the evaluation of our model in two parts. First, we discuss the performance on veterinary data compared to other methods. Second, we analyse the different components and strategies of our approach.

\subsection{Performance Comparison}
\begin{figure*}[ht]
  \vskip 0.2in
  \begin{center}
    \centerline{\includegraphics[width=1.0\columnwidth]{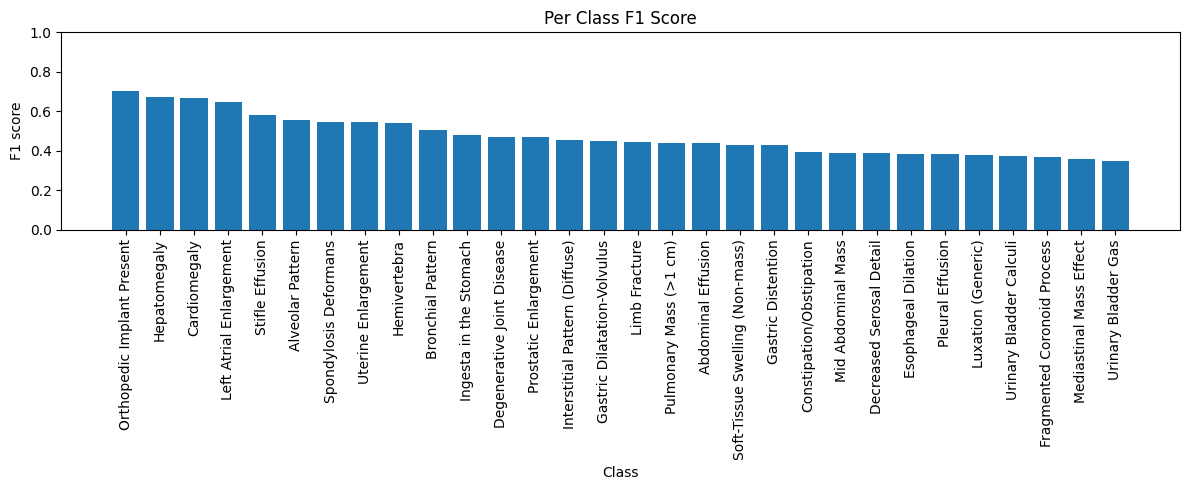}}
    \caption{F1 scores for the 30 most common conditions when using our V-REX model.}
    \label{fig:mostcommon}
  \end{center}
\end{figure*}

We provide evidence for the performance of our model by comparing it to powerful generalists (exemplary, OpenAI o3 \cite{openai2025o3}) and a fine-tuned VLMs (exemplary, PaliGemma \cite{paligemma}). To compare results, we let a model generate a report, then feed both the ground truth report written by a veterinarian and the generated report into o3. We prompt the model to first extract a list of existing findings according to a fixed catalogue from the ground truth, then do the same to the generated report, measuring the global F1 score. We exhibit results showing a strong advantage of our approach in \cref{tab:model_comparison}, despite our model only having a fraction of the parameters. Note that fine-tuned versions of PaliGemma do not need to learn to speak first, giving our model a singificant advantage in the first days of training. After about 4 days, the losses of PaliGemma variants stopped improving.\\
We conclude that the existing expert knowledge, feature extraction, and the language knowledge in a generalist VLM is still below what we can achieve with a model orders of magnitudes smaller if it is crafted properly, and that even fine-tuning an existing model is not better than training an expert from scratch with out recipe. We provide a more detailed evaluation for scores on different conditions in \cref{fig:mostcommon}.
\begin{table}[t]
  \caption{We measure training time of using one H100 on our dataset, with PaliGemma being fine-tuned to our data. Note that these are micro F1 scores (globally across all classes of findings). Both LorA and Finetuning for Paligemma did not improve on their train loss anymore.}
  \label{tab:model_comparison}
  \begin{center}
    \begin{small}
      \begin{sc}
        \begin{tabular}{lccc}
          \toprule
          Model & Days & Trainable & F1 $\uparrow$ \\
           & on H100 & Parameters & Score \\
          \midrule
          \midrule
          o3 & unknown & unknown & $0.2$ \\
          PaliGemma, LorA & $1$ & $3.3$M & $0.11$ \\
          PaliGemma, LorA & $5$ & $3.3$M & $0.13$ \\
          PaliGemma, Finetuning & $1$ & $3$B & $0.34$ \\
          PaliGemma, Finetuning & $5$ & $3$B & $0.38$ \\
          ours, light & $1$ & $504$M & $0.34$ \\
          ours, light & $5$ & $504$M & $0.38$ \\
          ours, light & $7$ & $504$M & $0.39$ \\
          ours, big & $~600$ & $925$M & $0.4$ \\ 
          \bottomrule
        \end{tabular}
      \end{sc}
    \end{small}
  \end{center}
  \vskip -0.1in
\end{table}

\subsection{Analysis}
\textbf{Data Efficiency}
\begin{table}[t]
  \caption{Top-1 accuracy for next text token prediction on the test set after 30M samples seen, across different dataset sizes.}\label{tab:data_efficiency}
  \begin{center}
    \begin{small}
      \begin{sc}
        \begin{tabular}{lcccc}
          \toprule
          Dataset size & $0.1\,\text{M}$ & $0.5\,\text{M}$ & $3\,\text{M}$ & $15\,\text{M}$ \\
          \midrule
          Accuracy (top-1) $\uparrow$ & $0.210$ & $0.327$ & $0.351$ & $\textbf{0.394}$ \\
          \bottomrule
        \end{tabular}
      \end{sc}
    \end{small}
  \end{center}
  \vskip -0.1in
\end{table}
A known issue when training VLMs is data scarcity. Hence, for veterinary data, we evaluate how much data is actually needed to perform well. For simplicity, we hence measure performance on different dataset sizes for the same amount of steps \cref{tab:data_efficiency}. The diminishing results show that a fairly capable model can be trained with as little as $3m$ text-image pairs.\\ We also measure the effects of using our existing data more effective through our proposed generative pre-training on images in \cref{tab:pre_training}.
\begin{table}[t]
  \caption{Generative pre-training effect, measure as the best value occurring on the $3m$ subset. We pre-train for 15 epochs, giving the number of cluster centres used for discretisation to allow autoregressive pre-training. 'None' shows without pre-training.}\label{tab:pre_training}
  \label{sample-table}
  \begin{center}
    \begin{small}
      \begin{sc}
        \begin{tabular}{lcccr}
          \toprule
          Measure & None & $\textbf{512}$ & $4096$ & $32768$ \\
          \midrule
		  Accuracy (top-1) $\uparrow$ & $.3769$ & $\textbf{.3898}$ & $.3806$ & $.3753$\\
		  Accuracy (top-5)$ \uparrow$ & $.6739$ & $\textbf{.6857}$ & $.6741$ & $.6694$\\
		  Perplexity $\downarrow$ & $14.81$ & $\textbf{13.85}$ & $15.48$ & $15.26$\\
          \bottomrule
        \end{tabular}
      \end{sc}
    \end{small}
  \end{center}
  \vskip -0.1in
\end{table}~\\

\subsection{Grounding Through IFG} 
\begin{figure}[ht]
  \begin{center}
    \centerline{\includegraphics[width=1.0\columnwidth]{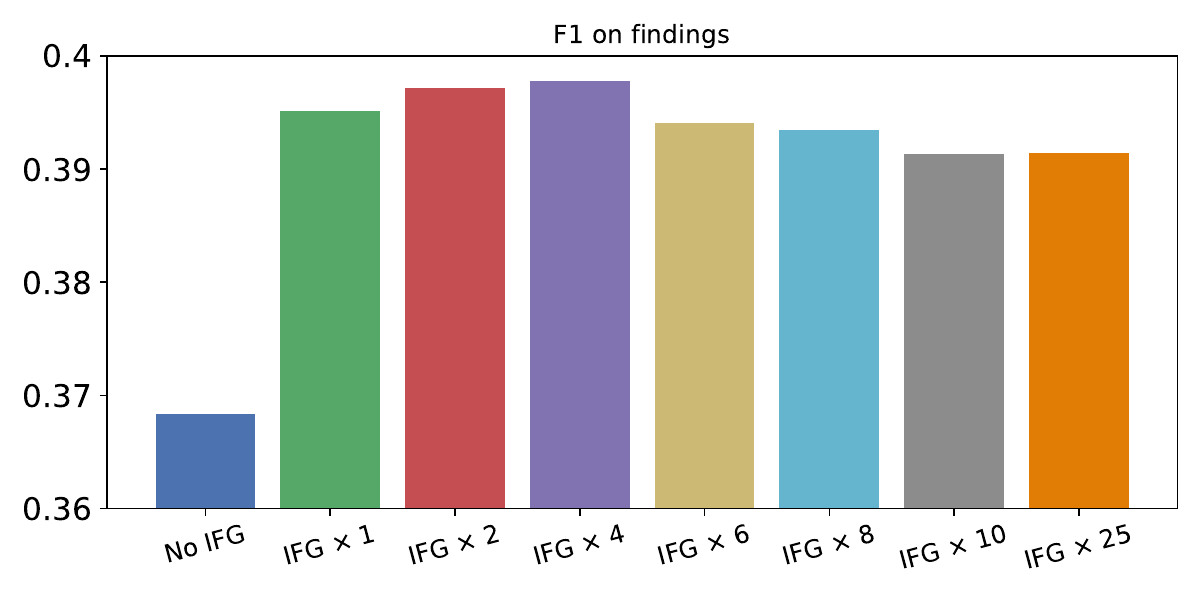}}
    \caption{We show the effectiveness of our additional grounding through comparing performance with and without IFG.}
    \label{fig:effectiveness_ifg}
  \end{center}
\end{figure}

We evaluate the effectiveness of our post-training Image Free Guidance in \ref{fig:effectiveness_ifg}, seeing a noticeable increase in correct reports through amplifying image impact on token probabilities. Especially on smaller datasets, this technique can help amplify the models capabilities without more training, which is prone to overfitting on smaller datasets.

\section{Conclusion}
We introduce a recipe to train a small expert VLM from scratch. We conclude that better data usage through pre-training, additional grounding for better results, a strong pre-trained feature extractor like RAPTOR, and an expert-level tokenisation are key components to obtain a cheap, lightweight model from scratch. Our results can help to refute the prejudice that fine-tuning a massive, general expert is the only viable approach: A small, lightweight model with the right design choices can be even more effective than the general purpose behemoths.

\bibliographystyle{splncs04}
\bibliography{main}

\newpage
\appendix

\section{Binary Subspace Quantisation}\label{app:bin_subspace_quant}
When applying k-means to divide an input space into k different clusters that we can use for generative autoregressive pre-training, memory use can skyrocket when assigning our continuos vectors of our input space of a vector $x$ with $d$ dimensions: For a number of $k$ cluster centres, we need to compute a $d \times k$-dimensional matrix per vector to find the closest cluster centre.\\
Our idea is to simply round each dimension of an input vector $x$ in such a way that all possible value combinations of the dimensions form our $k$-many different indices. In the simplest form, we can just subtract the average $x$-vector from an input, then check for each dimension whether it is great or equal zero, or below. This gives us $2^d$-many different quantised vectors. To have control over the number of quantised vectors, we project $x$ to a dimensionality of $n$ first using a projection $A$ that we optimise to be as lossless as possible.\\~\\
More formally, let $\mathcal{X} = \{x_i\}_{i=1}^N$, $x_i \in \mathbb{R}^d$, be a dataset of $d$-dimensional vectors. We seek a low-dimensional representation and a discrete clustering induced by binary codes.
We learn an encoder matrix $A \in \mathbb{R}^{n \times d}$ and a decoder matrix
$A' \in \mathbb{R}^{d \times n}$ by minimizing the reconstruction error

\begin{equation}
\min_{A, A'} \; \sum_{i=1}^N \left\| A' A x_i - x_i \right\|_2^2 .
\end{equation}

This corresponds to learning a rank-$n$ linear subspace approximation of the data.
Each data point is then projected into a lower dimensional latent space:

\begin{equation}
z_i = A x_i \in \mathbb{R}^n .
\end{equation}

We compute the empirical mean of the latent representations

\begin{equation}
\mu = \frac{1}{N} \sum_{i=1}^N z_i ,
\end{equation}

and centre the latent vectors:

\begin{equation}
\tilde{z}_i = z_i - \mu .
\end{equation}

We then assign a binary code $b_i \in \{-1, +1\}^n$ to each centred latent vector by applying an element-wise sign function:

\begin{equation}
b_i = \operatorname{sign}(\tilde{z}_i),
\end{equation}

where
\[
\operatorname{sign}(u)_j =
\begin{cases}
+1, & u_j \ge 0, \\
-1, & u_j < 0 .
\end{cases}
\]

Each binary vector $b_i$ then corresponds to one of $2^n$ possible binary codes.
The cluster assignment function is therefore

\begin{equation}
c_i = b_i \in \{-1, +1\}^n,
\end{equation}

implicitly partitioning the latent space into $2^n$ axis-aligned regions in the projection space.
\paragraph{Complexity Example}
We consider clustering $d = 128$ dimensional vectors into $K = 4096$ clusters. In standard $K$-means clustering, each data point is assigned to the nearest cluster center by computing the distance to all $K$ centroids. This requires $K \times d = 4096 \times 128 = 524{,}288$ scalar multiplications per data point, ignoring lower-order terms.\\
In our method, the number of clusters is determined by the number of latent dimensions $n$, with $2^n = 4096 \;\;\Rightarrow\;\; n = 12$. Each data point is projected into a $12$-dimensional latent space via a linear mapping $z = A x, \quad A \in \mathbb{R}^{12 \times 128},$ which only requires $n \times d = 12 \times 128 = 1{,}536$ scalar multiplications per data point. The cluster assignment is then obtained by applying an element-wise sign function to $z - \mu$, which incurs negligible additional cost.\\
The proposed method reduces the per-point cluster assignment cost by over two orders of magnitude while implicitly representing the same number of clusters, although at the cost of only allowing axis-aligned boundaries in the projection space.



\section{Unsuccessful Approaches and Negative Results}\label{app:unsuccessful}
As our sequences are somewhat structured and have a strong relation to each other, we explored \textbf{Multi-Token Prediction} \cite{multitoken}. We further hoped to obtain more training signal per token, as training this way not only gives a single signal (the next token), but the next few tokens. Our hopes were particularly on getting a higher data utilisation from this. However, in all our experiments, performance was always slightly worse or at best, roughly even. As multi-token prediction usually requires multiple orders of magnitude bigger models than ours to be effective, producing similar results does indicate that for a model that is an order of magnitude bigger, multi-token prediction could be beneficial.\\
Input and output \textbf{embedding tying} of the text does not help. From our experiments, we conclude that the model size (about $500\text{M}$ parameters), compared to the vocabulary size of about $32{,}000$ tokens, suffers more from the restriction of using shared weights than it benefits from the reduced parameter count and improved alignment between input and output embeddings. We also speculate that, since the decoded tokens stem from a representation that has been enriched with image information, their modality changes enough to negatively impact performance when the embeddings are tied together.\\
We further experimented with \textbf{random augmentations} of the input image embeddings and different \textbf{optimiser $\beta$-values}, all to no avail.\\
Lastly, we made the important discovery that putting a model with dropout layers in \textbf{evaluation mode} to avoid dropping 10 percent of the information per layer actually helps performance (unsurprisingly...).

\end{document}